\documentclass[letterpaper, 10pt, conference]{ieeeconf}
\usepackage{blindtext, graphicx}
\graphicspath{{pictures/}}

\usepackage{subfigure}
\usepackage{wrapfig}

\IEEEoverridecommandlockouts

\begin{document}
\title{Open World Assistive Grasping Using Laser Selection}

\author{Marcus Gualtieri$^{1}$, James Kuczynski$^{2}$, Abraham M. Shultz$^{2}$, Andreas ten Pas$^{1}$, Holly Yanco$^{2}$, Robert Platt$^{1}$% <-this % stops a space
\thanks{$^{1}$College of Computer and Information Science, Northeastern University, Boston, Massachusetts
        {\tt\small atp@ccs.neu.edu}}%
\thanks{$^{2}$Department of Computer Science, University of Massachusetts Lowell, Lowell, Massachusetts
        {\tt\small ashultz@cs.uml.edu}}%
\thanks{This work has been supported in part by the National Science Foundation through IIS-1426968 and IIS-1427081, NASA through NNX16AC48A and NNX13AQ85G, and ONR through N000141410047.}
}

% make the title area
\maketitle

\begin{abstract}
%\boldmath
Many people with motor disabilities are unable to complete activities of daily living (ADLs) without assistance. This paper describes a complete robotic system developed to provide mobile grasping assistance for ADLs. The system is comprised of a robot arm from a Rethink Robotics Baxter robot mounted to an assistive mobility device, a control system for that arm, and a user interface with a variety of access methods for selecting desired objects. The system uses grasp detection to allow previously unseen objects to be picked up by the system. The grasp detection algorithms also allow for objects to be grasped in cluttered environments. We evaluate our system in a number of experiments on a large variety of objects. Overall, we achieve an object selection success rate of 88\% and a grasp detection success rate of 90\% in a non-mobile scenario, and success rates of 89\% and 72\% in a mobile scenario.
% Results of the system's performance with a variety of objects are presented.
\end{abstract}

\IEEEpeerreviewmaketitle

% ========================================================================================================================
\section{Introduction}

There are millions of individuals in the United States who have motor disabilities that make it difficult to grasp or manipulate common household objects~\cite{brault2012americans}. 
However, there are very few options for using robotic manipulation technologies to help these people perform the manipulation tasks required by activities of daily living (ADLs). 
There are three key challenges in developing assistive manipulators for people with motor disabilities. 
First, robotic arms that might be deployed in the home are typically expensive. 
For example, the Jaco assistive arm costs more than \$40K for a single arm. 
Second, it can be very challenging to teleoperate these arms. 
Direct teleoperation requires constant mental effort and attention, and it is a particular challenge for users who have both motor and cognitive impairments. 
Finally, each potential user has a unique set of abilities and challenges, requiring the interface to an assistive system to have an adaptable set of interaction methods. 
This paper describes a system that addresses each of these challenges.

We have developed a system that integrates a robotic arm/hand controlled by a laser-directed grasping system with an assistive scooter. To address the cost issue, we used an arm from a Rethink Robotics Baxter robot. At \$25K for the entire two-armed robot, the cost of a single arm is substantially lower than other commercially available robot arms, whether intended for the assistive technology market or other domains.

\begin{figure}
    \centering
    \includegraphics[width=\columnwidth]{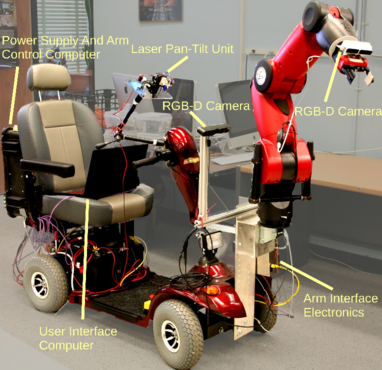}
    \caption{We have developed an assistive grasping system that can help people with motor disabilities to perform the activities of daily living. It is comprised of a single arm from a Rethink Robotics Baxter robot that has been mounted to the front of a Golden Avenger mobility scooter. Novel user interface and grasping capabilities enable this system to automatically grasp objects that the user points to using a laser pointer.}
    \label{img:scooter}
\end{figure}

The system uses grasp detection in order to provide assistance with picking up objects in the home during ADLs. 
Typical approaches to robotic grasping generally require the robot to have access to a known mesh model of the object to be 
grasped (e.g.,~\cite{klank2009,chitta2012,glover2013}). 
However, it is clearly impractical to require users of the system to obtain these models in order to use the system. 
Further, the use of models can limit the system's ability to grasp novel objects, thus ``closing'' the world that is available for manipulation. 
By finding grasps in sensor data, the robot can operate in an ``open'' world, where the objects it can grasp are not limited {\em a priori} by the models available to the system. 
Grasp detection finds locations in the environment where the robot can safely move its hand and expect to grasp something when the fingers are closed. 
By combining automatic grasp detection with a customizable laser pointer selection system and an inexpensive arm, we create a system capable of autonomously grasping anything that the robot hand is physically capable of holding.

\section{System Overview}

Our system is composed of one robotic arm from a Rethink Robotics 
Baxter robot mounted on a Golden Avenger mobility scooter as shown 
in Figure~\ref{img:scooter}. The Baxter arm is a 7-DOF arm with a 
1-DOF (parallel) gripper. 
The only 
difference between our ``dismembered'' Baxter arm and a standard Baxter arm 
is the fact that it has been separated from the rest of the body.\footnote{In order to use both of the Baxter's arms individually, a second control computer must be purchased from Rethink Robotics. Some software configuration changes are also necessary to operate the arm without the head, torso, or other arm typically found on a Baxter robot.} 
The mobility scooter is COTS equipment designed for users 
with good upper-body mobility, but limited strength and endurance.
The scooter cannot drive itself, but autonomous mobile manipulators have been explored by e.g. Choi \emph{et al.} \cite{choi2008laser}.
To make the arm mobile, a commercially available 1000W pure-sine AC inverter, produced by Samlex, has been used to convert power from the 24v batteries in the scooter for the Baxter arm and the control computer for the arm. 

In our system, the 
Baxter robot, the laser pointer based user interface, and the grasp control 
subsystems are all implemented as ROS nodes on computers located 
on board the scooter.
The basic operations of the arm are controlled by the PC that is typically built into Baxter robots, running the same controllers and ROS nodes that are typical in standard Baxter robots. 
The grasp selection, laser detection, and kinematic control nodes are run on a separate laptop, connected directly to the Baxter PC with a network cable. 
This laptop can run for approximately 1.5 hours on its internal battery, but it can also be powered from the scooter-mounted inverter for longer runs. 
Adding the arm and laptop to the power draw on the scooter batteries will reduce the amount of time that the scooter can be operated between charges, but the system as a whole can operate for multiple hours without recharging. During manipulation experiments, with the scooter repeatedly driving small distances and powering both the laptop and the arm, the system was run for approximately 5 hours, in which it used less than half of the available charge from the scooter battery, as indicated by the scooter's built-in power display. 

\section{User Interface}

The goal of the laser pointer based user interface is to enable people with motor and/or cognitive disabilities to operate the grasping system. 
We want these users to be able to indicate easily which objects in the surrounding environment are to be grasped. 
Since this system is to operate in open environments, it is important not to require prior models of the objects to be grasped. 
Ideally, the system would be able to grasp anything that the user points at, regardless of exactly what it is, although the system will be limited by the capability of the installed gripper. 

\begin{figure}
    \vspace{1mm}
    \centering
    \subfigure[]{\includegraphics[height=1.2in]{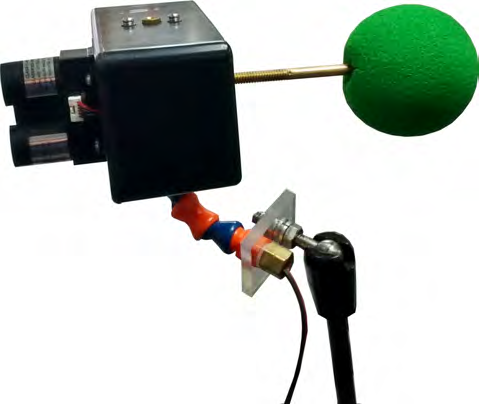}}
    \subfigure[]{\includegraphics[height=1.2in]{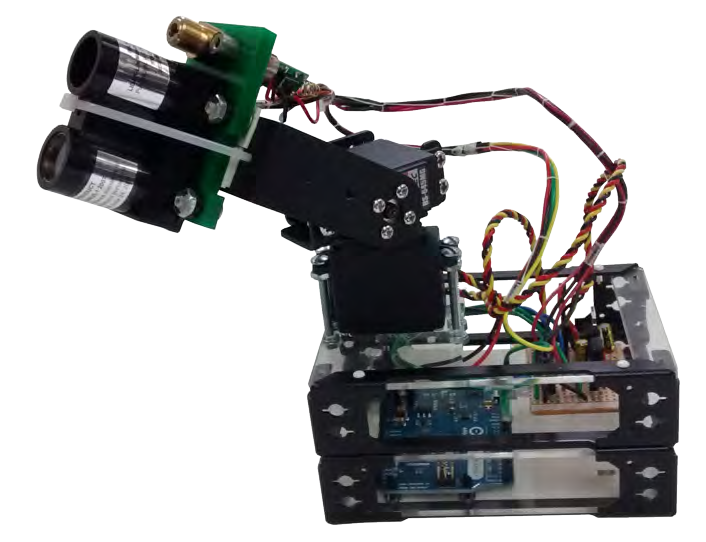}}\\
    \subfigure[]{\includegraphics[height=1.4in]{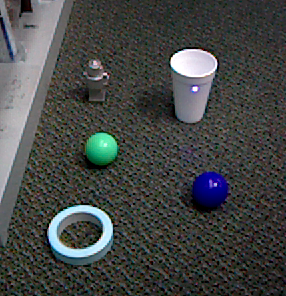}}
    \hspace{0.5cm}
    \subfigure[]{\includegraphics[height=1.4in]{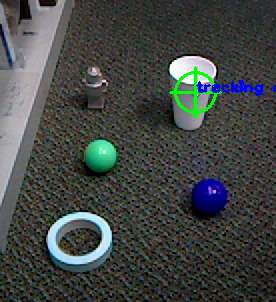}}
    \caption{(a) Manual laser interface. The foam ball can be replaced by other hand grips for individual user's needs. (b) Servo laser interface. This device can be controlled by a variety of computer access methods. 
    (c) Laser point in a scene. (d) Detection of the laser point.}
    \label{fig:laserpointer}
\end{figure}

The laser-grasp idea is a good one for control of a grasping system, but the interface must take into account the ability of the user.
Our general approach is the same as that of Kemp {\em et al.} who used a laser pointer to enable users to indicate objects in an environment that should be grasped by a robot~\cite{kemp2008point}. 
Later extensions by Choi {\em et al.} used an ear-mounted laser for patients with limited upper-limb motor control, but good head movement, which extends the utility of their system to people with severe motor disabilities \cite{choi2008laser}.

We developed two different interfaces to the laser pointing system. 
The first is a manual interface, designed for users with tremors affecting their hand and arm motion.
The laser pointer is mounted on a frictional ball joint. The user points the laser by manipulating a 
foam ball (see Figure~\ref{fig:laserpointer}(a)). The second interface is designed for users who have more severe upper body impairments. The laser 
pointers are mounted on a pan-tilt system controlled by a pair of servos (Figure~\ref{fig:laserpointer}(b)). The servos allow us to control the position of the laser using a joystick or mouse, and could be controlled by sip-and-puff or single-switch scanning to accommodate users with quadriplegia. 

The user interfaces are mounted to an articulated arm which is in turn mounted to the handlebars of the scooter. 
The articulated arm permits the interface to be placed where the user can operate it, and where the path of the laser is minimally obstructed by the arm and other hardware. 

The interface 
itself contains two (5mW 650nm) lasers, one red and one 
green, that are used indicate whether a point is within the 
workspace of the robot end effector or not. When the point is 
outside the workspace of the robot arm (we approximate this 
as within a 2 meter envelope of the laser), the interface illuminates it with 
the red laser. When it is within the manipulator workspace, it is 
illuminated with the green laser.\footnote{We are exploring the use of shape as well as color, to assist people who have red-green color blindness.}

The system detects the position of the point indicated by the laser by 
pulsing the laser at 5 Hz and looking for differences in successive 
camera frames. The area in front of the robot is sensed by a PrimeSense Carmine sensor mounted near the base of the Baxter arm (labeled as ``RGB-D Camera'' in Figure~\ref 
{img:scooter}). Areas of interest are calculated by differencing 
successive frames and looking for large changes in intensity, as well 
as areas of high brightness after automatic exposure correction based on 
\cite{mevsko2013laser}. Since 
large differences can be caused by motion in the environment, the system filters the resulting regions based on size 
and color. Rejecting detections that are not persistent over several frames further reduces the chance of false detections. Figure~\ref{fig:laserpointer}(c-d) shows an 
example of a laser detection. After correctly pointing the laser, the user 
indicates to the system to go ahead with the grasp and the detected 
point is provided to the grasp system.

\section{Using Active Sensing to View the Object to be Grasped}

\begin{figure}
    \centering
\vspace{3 mm}
\subfigure[]{\includegraphics[height=1.6in]{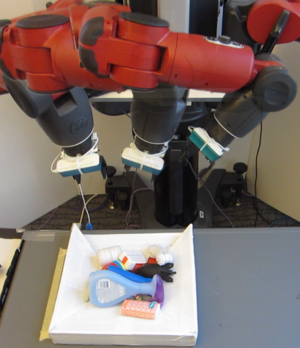}}
    \subfigure[]{\includegraphics[height=1.6in]{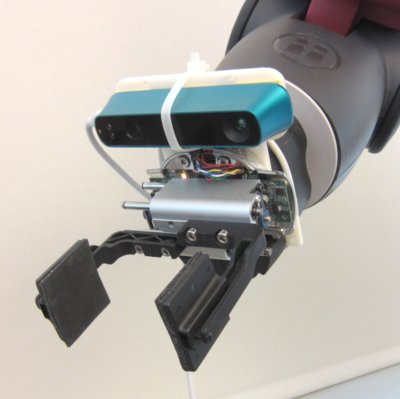}}
  \caption{(a) In our prior work, we used a SLAM package to model the 
  scene by moving the depth sensor in an arc around the objects of 
  interest. (b) Structure sensor mounted on wrist for getting detailed scans of target objects. Grippers were equipped with stock Baxter square pads with foam added to the surfaces.}
\label{fig:clutter_sensor_motion}
\end{figure}

%\begin{figure}
%\begin{center}
%    \includegraphics[height=1.5in,trim={0.4in 0.1in 0.6in 
%    0.5in},clip]{pictures/wrist_mounted_depthsensor.png}
%\end{center}
%  \caption{Structure sensor mounted on wrist for getting detailed scans of target objects. Grippers were equipped with stock Baxter square pads with foam added to the surfaces.}
%  \label{fig:gripperAndSensor}
%\end{figure}

The grasp detection system uses a point cloud that it takes as input 
to calculate predicted grasp configurations, i.e., grasp detections. 
The quality of this point 
cloud has a significant bearing on the performance of grasp 
detection. In particular, our prior work has shown that the same 
grasp detection-based system achieves a 9\% higher grasp success rate 
(93\% instead of 84\% grasping in dense clutter) when using a point 
cloud obtained using dense metric SLAM instead of a point cloud 
obtained directly from the depth sensors~\cite{Gualtieri2016}. This result suggested that a similar approach would improve the accuracy of the scooter system's grasping. %However, use of SLAM implies 
%that a depth sensor exists that will be moved through the 
%environment in order to view the scene from multiple perspectives. 
In our prior work, we mounted the depth sensor near the robotic hand 
and moved the hand through a pre-programmed trajectory around the 
objects of interest (Figure~\ref{fig:clutter_sensor_motion}). 
Unfortunately, that strategy cannot be used in the current situation 
because the objects to be grasped can be found in different places 
around the workspace, e.g. on shelves, tables, {\em etc.}. 
Futhermore, the user could drive the scooter into a location where the pre-programmed trajectory would collide with objects in the environment, such as a narrow pantry or closet entrance. 
To avoid this problem, we gave the system the ability to plan a sensor trajectory that avoids obstacles while scanning the relevant portion of the workspace.

We refer to the trajectory that the sensor takes in order to 
observe the object to be grasped as the ``information gathering 
trajectory''. Successfully planning this trajectory requires 
several constraints to be met. First, the view of the point of 
interest must not be occluded by obstacles in the environment. 
Second, it is essential to select viewpoints beyond the minimum 
range of the depth sensor. 
%All depth sensors have a minimum range. 
%In our case, since the Structure IO sensor has a 40 cm minimum range, 
%which excludes a significant amount of the workspace. 
Finally, it 
is important to minimize the length of the information gathering 
trajectory. SLAM packages (we use InfiniTAM~\cite{prisacariu2014framework} in this work) 
can lose track of the environment relative to the voxelized grid 
that is created when there is not enough structure in the viewed 
environment to enable the matching algorithm (e.g., iterative closest 
point~\cite{besl1992method}) to function robustly. This is a 
particular risk at close range because there is often less structure 
than in room-scale scenarios. We reduce this risk by minimizing trajectory length.

Given the above constraints, the system plans the information gathering trajectory as follows. First, it 
is assumed that the system is given a point of interest -- in our 
case, this will be the laser point. 
The system constructs a sphere about the point of interest with a radius of 42cm,  
slightly larger than the 40cm minimum range of the Structure sensor. 
Pairs of points are then sampled uniformly at 
random from the sphere until a pair is found such that the points 
are less than 22 cm apart and have collision-free inverse kinematic (IK) 
solutions that permit the sensor to point directly at the point of interest. 
IKFast is used to generate the IK solutions and 
OpenRAVE is used to test for collisions with known obstacles, such as parts of the scooter. Once a feasible pair of points is 
found, TrajOpt~\cite{Schulman2013} is used to generate a trajectory 
that is collision-free with respect to the PrimeSense point cloud, remains outside the minimum range of the 
depth sensor, causes the sensor to point directly toward the point 
of interest throughout, and travels between the two sampled points. 

\section{Grasp Detection}

Grasp detection is a key element of our system because it enables us 
to grasp novel or unmodeled objects. Traditional approaches to 
robotic grasping are based on localizing and grasping specific known 
objects~\cite{klank2009,chitta2012}, or grasping based on behaviors that take advantage of common features of environments built for humans \cite{jain2010assistive}.
While these methods work well 
in structured environments, or where object geometries are known ahead 
of time, they are less well suited for unstructured or cluttered real world 
environments. In contrast, grasp detection methods search for good 
grasp configurations directly in sensor data -- in a point cloud or 
an RGB-D image~\cite 
{lenz_rss2013,redmon2015real,fischinger2013learning, 
detry2013a,herzog_icra2012,kappler2015leveraging,tenpas_isrr2015,Gualtieri2016}. 
These grasps are full 6-DOF poses of a robotic hand 
from which a grasp is expected to be successful. Importantly, there 
is no notion of ``object'' here. A detected grasp is simply a hand 
configuration from which it is expected that the fingers will 
establish a grasp when they are closed. Grasp detection methods are particularly well suited to open world environments because they do not require mesh models.

\begin{figure}
\vspace{3mm}
    \centering
\subfigure[]{\includegraphics[height=1.2in]{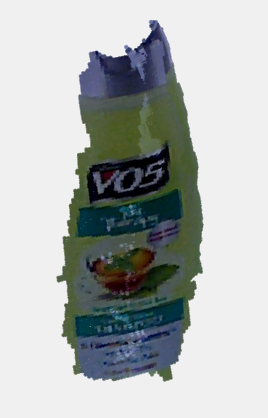}}
\hspace{0.2cm}
\subfigure[]{\includegraphics[height=1.2in]{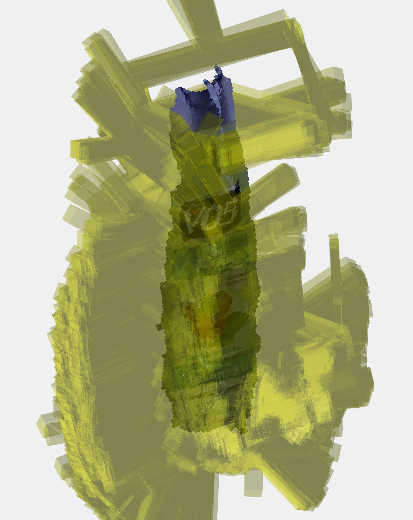}}
\hspace{0.2cm}
\subfigure[]{\includegraphics[height=1.2in]{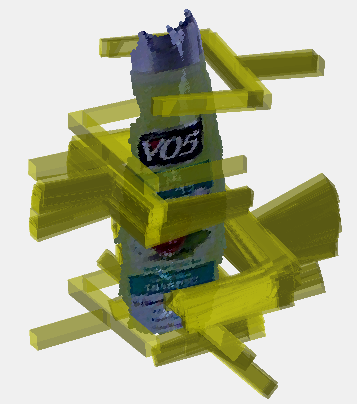}}
  \caption{(a) Input point cloud to the grasp detection system; (b) Grasp candidates that denote potential grasp configurations; (c) High scoring grasps.}
\label{fig:gpd_example}
\end{figure}

Our grasp detection algorithm has the following two steps: grasp 
candidate generation and grasp scoring. During candidate generation, 
the algorithm hypothesizes a large number of potential grasps, 
generated by searching for hand configurations that are obstacle 
free with respect to the point cloud and contain some portion 
of the cloud between the two fingers. For more detail on this process, 
see \cite{tenpas_isrr2015}. While some of these hand 
configurations may turn out to be grasps, most of them will not. 
Figure~\ref{fig:gpd_example}(b) shows an example of grasp candidate 
generation for a point cloud of a shampoo bottle. This phase of the 
process serves only to focus attention of the algorithm on promising 
candidates. The next step is to score the grasp candidates according 
to how likely they are to be true grasps. Here, some form of machine 
learning is typically used to rank the candidates. In our case, the system uses a 
four-layer-deep convolutional neural network to make grasp 
predictions based on projections of the portion of the point cloud 
contained between the fingers. Figure~\ref{fig:gpd_example}(c) 
shows the grasps that were scored very highly by our system. More 
details on our method can be found in~\cite{Gualtieri2016}.

\section{Grasp Selection}

If grasp detection runs successfully, it typically finds hundreds of 
hand configurations that are likely to be good grasps given the 
gripper and local object geometry. The system filters those grasps as 
follows. First, since the user wants to grasp the object that was 
illuminated by the laser pointer, the system eliminates all grasps more than 
6cm from the laser point. Next, the system filters 
out all grasps that do not have collision-free IK solutions. A key 
drawback of grasp detection in its current form is that it only 
takes into account the geometry of the object surfaces to be 
grasped. It ignores other relevant information such as distance to 
the center of gravity of an object, whether an object is near the 
top of a pile or not, and how heavy the object is likely to be. 
Currently, we compensate for this by encoding simple 
heuristics into a cost function that is used to select grasps. The 
cost function gives preference to grasps with large $z$ coordinate 
values that are likely to be near the top of a pile of objects. It 
also gives preference to grasps that most closely align with the 
gravity vector. We call these grasps ``top grasps'' because the gripper approaches the object from the top down, as opposed to ``side grasps'', where the gripper approaches the object from the side of the object. We have empirically found that top grasps
succeed more often, and have an advantage for avoiding collisions with surrounding objects. However, the top grasp heuristic is not used if the grasp position is 
less than 34cm beneath an obstacle in the environment. This allows the 
robot arm to reach objects on a shelf that would not be reachable with top grasps because the robot arm will not fit between the object and the shelf above it. 
The cost function penalizes grasps that would require the manipulator to 
travel near its joint limits or which can only be grasped by 
closing/opening the gripper fingers to their limits. Finally, the 
cost function encodes a preference for grasps that can be reached by 
moving the arm the shortest distance in configuration space from its 
final viewing configuration. The grasp with the lowest cost is 
selected for execution, and TrajOpt is used to generate the motion 
plan for the grasp and for bringing the arm safely back to the 
basket to deposit the object.

\section{Experiments}

We evaluated our system using two types of experiments. First, we 
performed a tabletop experiment where we evaluated the grasping 
system in isolation from the rest of the system. Then we performed an 
experiment that evaluated system performance in scenarios close to our 
targeted use-case scenarios.

\subsection{Evaluating Grasping in Isolation}
\label{sect:isolationexp}
\begin{figure}
\vspace{3mm}
\begin{center}
    \includegraphics[height=2.0in]{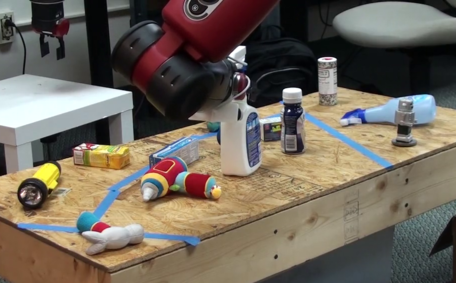}
\end{center}
  \caption{Experimental scenario for evaluating the grasping system by itself.}
  \label{fig:tablegrasp}
\end{figure}

Since grasping is the core functionality of this system, it is 
important to characterize grasp performance as accurately as 
possible. In the standard configuration where Baxter is mounted on its 
stock stand, it is possible to achieve a 93\% grasp 
success rate for novel objects presented in dense clutter such as 
that shown in Figure~\ref{fig:clutter_sensor_motion}(a)~\cite{Gualtieri2016}. The 
current scenario is a little different because we are now 
identifying the object to be grasped using the laser. Nevertheless, 
we should be able to achieve a similar grasp success rate in the 
current scooter configuration. To test this, we performed a series 
of 15 trials in the tabletop scenario shown in Figure~\ref 
{fig:tablegrasp}. At the beginning of each trial, six objects were randomly selected from a set of 30 possible objects (shown in Figure~\ref{fig:roomlayout}(c)) and 
placed on the table within the workspace of the Baxter arm. Then, 
the system was directed to each of the 
six objects in a randomly determined order. 
If the selected grasp belonged to the object that 
the user targeted, then this was considered an object detection 
success. Otherwise, it was an object detection failure. Similarly, we 
measured the grasp success rate. Once a grasp was selected for 
execution (whether or not that grasp was on the target object), we tracked whether the robot was able to 
successfully grasp, transport, and drop the object into a 
container. If the robot was successful in performing all of 
these steps, then the grasp was considered a success. Otherwise, it 
was considered a failure. 

Overall, we performed 15 trials with 6 objects on each trial (a total 
of 90 objects for detecting/grasping). There were 123 
attempts to detect an object using the system. (This 
number exceeds 90 because detection and grasp errors 
caused us to attempt detection multiple times for some 
objects.) Only 108 detections were successful out of the 123 
attempts, giving us an 87.8\% detection success rate. Out of the 90 
objects, there were 87 grasp attempts. (The remaining objects 
rolled or fell out of the workspace due to direct or indirect 
collisions caused by the manipulator.) Out of the 87 grasp attempts, 
78 were successful, giving us an 89.6\% grasp success rate. As a 
whole, we consider these to be good results. The 89.6\% grasp 
success rate is close to what we have obtained in the lab under 
ideal conditions. The 87.8\% object detection success rate is lower than desired, but it could likley be improved using standard segmentation strategies.

\subsection{Evaluating Grasping In-Situ}

\begin{figure}
\begin{center}
    \subfigure[]{\includegraphics[height=2.2in]{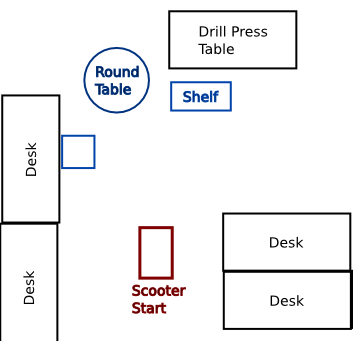}}
    \subfigure[]{\includegraphics[height=1.5in]{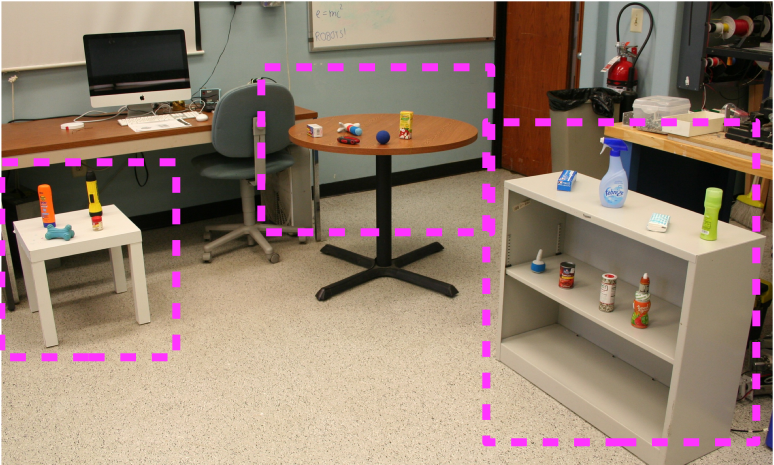}}
    \subfigure[]{\includegraphics[height=1.5in]{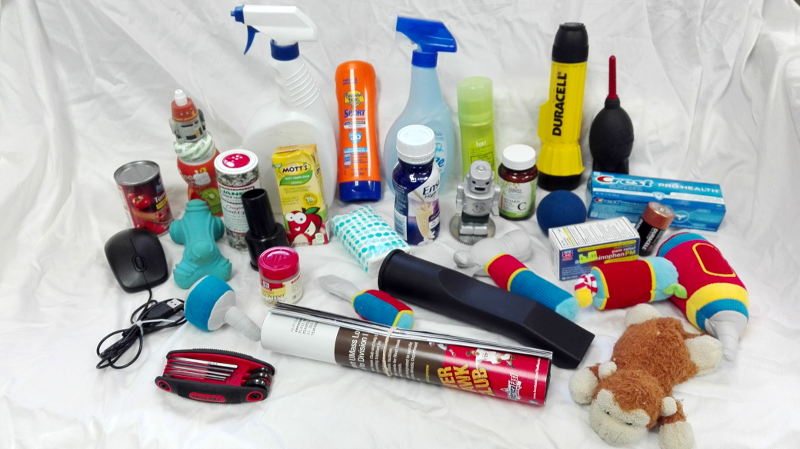}}
\end{center}
  \caption{(a,b) Room layout for in-situ experiments. (c) The 30 household 
  objects used in our experiments.}
  \label{fig:roomlayout}
\end{figure}

Our first priority is to to 
evaluate whether the proposed system can help people with 
motor and/or cognitive impairments to perform ADLs. As such, we performed an 
experiment to evaluate the system in a more realistic setting.
A user would sit in the scooter, drive it into a domestic 
environment, and attempt to grasp objects of interest. Ultimately, 
we plan to evaluate our system with patient populations at a 
collaborating rehabilitation facility. For this paper, we only evaluate the degree to which the system is capable of performing these tasks. Therefore, all of the current 
experiments were performed with able-bodied users.

The scooter was placed in a room with two tables and a shelf. Figure~\ref{fig:roomlayout} shows the 
layout of the room. The blue items in Figure~\ref{fig:roomlayout}(a) 
show the two tables and the shelf. These same pieces of furniture 
are shown in the magenta boxes in Figure~\ref{fig:roomlayout}(b). We 
performed a series of 5 trials. At the beginning of each trial, 10 
objects were randomly selected from the 30 total objects shown in 
Figure~\ref{fig:roomlayout}(c). These 30 objects were selected 
according to two criteria: 1) they represent some of the objects 
that might typically be used in daily life; and 2) the Baxter parallel 
gripper is mechanically capable of grasping them. We placed these 10 
objects on the three pieces of furniture in randomly assigned 
positions. Three objects were placed on each of the two tables and 
two objects were placed on each of the two shelves (the top shelf 
was 90 cm above the ground and the middle shelf was 60 cm above the 
ground). In total, there were $5 \times 10 = 50$ objects that 
could be grasped.

 \begin{figure}
 \vspace{3mm}
  \centering
  \includegraphics[height=1.8in]{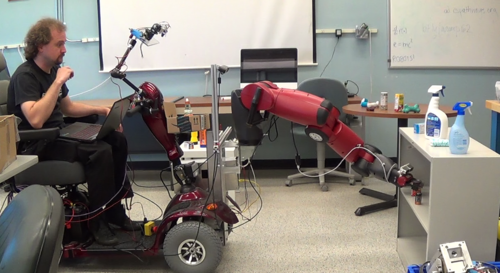}
  \caption{Scooter platform grasping an object in-situ.}
  \label{fig:abeonascooter}
 \end{figure}

Prior to grasping each object, the scooter was placed in the 
starting configuration as indicated in Figure~\ref 
{fig:roomlayout}(a). The arm was in a starting position chosen to minimize interference with the laser beam and user view of the workspace. After being instructed to grasp one particular 
object (the sequence in which objects were grasped was generated 
randomly), the user would drive the scooter up to the corresponding 
object, illuminate the object using the laser, and activate 
the automatic grasping process. Then the robot would scan the 
region around the laser point and attempt the grasp. 
We measured the object detection success rate and the grasp success rate as in the previous experiment. Figure~\ref{fig:abeonascooter} illustrates one of these grasps.

Out of the 50 objects available to be grasped, we attempted to 
detect an object 66 times (we attempted to grasp some objects 
multiple times because of grasp failures). Out of those 66 detection 
attempts, 7 failed giving us an 89.4\% object detection success 
rate. Note that this result is similar to the 87.8
\% detection success rate that was reported in Section~\ref
{sect:isolationexp} during the grasp experiments. Out of the 50 objects 
available to be grasped, there were a total of 61 grasp attempts, as some objects were attempted multiple times. Out of 61 attempts, there were 17 grasp failures, for a 72.1\% grasp success rate.  This grasp success rate is significantly lower than the
89.6\% success rate reported in Section~\ref{sect:isolationexp}. 
There were two primary failure modes that account for most of the 
additional failures. The first is collisions between the arm and the 
environment. In 4 out of the 17 grasp failures, the sensors failed 
to ``see'' important obstacles in the environment that the arm 
subsequently collided with. This failure mode did not occur in 
Section~\ref{sect:isolationexp} because the scooter was placed in a 
configuration that was largely obstacle-free. However, the main 
failure mode (13 out of the 17 grasp failures) was incorrect grasp 
detections. In these failures, the robot detected a grasp on the edge of
the table or shelf very close to the detected laser point. Since grasp detection does not detect objects {\em per 
se}, there is no intrinsic reason why it cannot detect a ``good'' 
grasp of the edge of a table or shelf.
Such failures did not occur 
in Section~\ref{sect:isolationexp} because the table 
location was known, and grasps on the table were excluded using workspace limits. 

We also evaluated the task in terms of the time required to drive the scooter up to an object and grasp it. Out of the 50 objects, the average time to grasp was 128s, the minimum was 44s, and the maximum was 374s. While this is a fairly large variation in time-to-completion, several of the trials were completed in close to the minimum time. Essentially, our system achieves the minimum time when each step of the grasp works as intended the first time -- when our SLAM package does not lose track, the arm does not collide with unseen objects in the environment, and the grasp works on the first try. If one of these elements fails, then the trial can still succeed -- it just takes longer. In order to put these times in context, we also performed a series of teleoperated grasp trials that were exactly the same as the automated trials, but where one of the researchers directly controlled the motion of the arm using a keyboard interface. The average time to complete a grasp in teleoperation mode was 40s. Even though this is less than the 44s minimum time taken by our automated system, it is close enough to suggest that our automated system could be a practical approach if we are able to correct for some of the failures described above. Moreover, it is important to note that even when the automated system takes longer to grasp than would be required by an able-bodied person to teleoperate the system, the automated capability is still extremely important for our target users, many of whom would find it challenging to teleoperate a robotic arm.

\section{Discussion and Future Work}

Overall, the results reported in this paper suggest that we may soon be able to begin user studies with target patient populations. The key thing to improve is the time required for a grasp. Waiting two minutes to grasp a single object could be frustrating to users. One way to speed up the process would be to eliminate the information gathering trajectory and instead rely on a point cloud from a single view. Perhaps with improvements in grasp detection or in choosing the correct viewing angle for the object, the same reliability can be achieved as with the SLAM method. Another major failure mode was arm collisions with unseen portions of the environment. In this case, we expect that adding additional depth sensors to the mobile base will eliminate most of these failures. Finally, we expect that improvements to the laser detection part of the system should reduce object selection failures.

\bibliographystyle{IEEEtran}
\bibliography{grabby}

% that's all folks
\end{document}